\documentclass[10pt,twocolumn,letterpaper]{article}

\usepackage{iccv}
\usepackage{times}
\usepackage{epsfig}
\usepackage{graphicx}
\usepackage{amsmath}
\usepackage{amssymb}
\usepackage{booktabs}
\usepackage{multirow}

\usepackage{enumitem}
\usepackage[square,numbers]{natbib}

\usepackage[pagebackref=true,breaklinks=true,letterpaper=true,colorlinks,bookmarks=false]{hyperref}

\iccvfinalcopy 

\def\iccvPaperID{9057} 

\ificcvfinal\fi

\newcommand{\Methodname}[0]{Seasonal Contrast}
\newcommand{\methodname}[0]{SeCo}

\begin{document}

\iftrue
\abovecaptionskip=1ex           
\belowcaptionskip=1ex           
\intextsep=0.1ex         
\textfloatsep=1.5ex      
\fi


\title{\Methodname{}:\\ Unsupervised Pre-Training from Uncurated Remote Sensing Data}

\author{Oscar Mañas$^{1,2}$ \quad Alexandre Lacoste$^1$ \quad Xavier Giró-i-Nieto$^2$ \quad David Vazquez$^1$ \quad Pau Rodriguez$^1$\\
\small $^1$Element AI \quad \quad $^2$Universitat Politècnica de Catalunya\\
{\tt\small oscmansan@gmail.com, pau.rodriguez@servicenow.com}
}

\maketitle
\ificcvfinal\thispagestyle{empty}\fi

\begin{abstract}
   Remote sensing and automatic earth monitoring are key to solve global-scale challenges such as disaster prevention, land use monitoring, or tackling climate change. Although there exist vast amounts of remote sensing data, most of it remains unlabeled and thus inaccessible for supervised learning algorithms. Transfer learning approaches can reduce the data requirements of deep learning algorithms.  However, most of these methods are pre-trained on ImageNet and their generalization to remote sensing imagery is not guaranteed due to the domain gap. In this work, we propose \Methodname{} (\methodname{}), an effective pipeline to leverage unlabeled data for in-domain pre-training of remote sensing representations. The \methodname{} pipeline is composed of two parts. First, a principled procedure to gather large-scale, unlabeled and uncurated remote sensing datasets containing images from multiple Earth locations at different timestamps. Second, a self-supervised algorithm that takes advantage of time and position invariance to learn transferable representations for remote sensing applications. We empirically show that models trained with \methodname{} achieve better performance than their ImageNet pre-trained counterparts and state-of-the-art self-supervised learning methods on multiple downstream tasks. The datasets and models in \methodname{} will be made public to facilitate transfer learning and enable rapid progress in remote sensing applications.\footnote{Code, datasets and pre-trained models are available at \url{https://github.com/ElementAI/seasonal-contrast}}
\end{abstract}

\section{Introduction}
Remote sensing is becoming increasingly important to many applications, including land use monitoring~\cite{foody2003remote}, precision agriculture~\cite{mulla2013twenty}, disaster prevention~\cite{schumann2018assisting}, wildfire detection~\cite{filipponi2019exploitation}, vector-borne disease surveillance~\cite{ippoliti2019defining}, and tackling climate change~\cite{rolnick2019tackling}. Combined with recent advances in deep learning and computer vision, there is enormous potential for monitoring global issues through the automated analysis of remote sensing and other geospatial data streams.

Remote sensing provides a vast supply of data. The number of Earth-observing satellites is continuously growing, with over 700 satellites currently in orbit generating terabytes of imagery data every day~\cite{neumann2019domain}. However, many downstream tasks of interest are constrained by a lack of annotations, which are particularly costly to obtain since they often require expert knowledge, or expensive ground sensors. In recent years, a number of techniques have been developed to mitigate the need for labeled data~\cite{laradji2020weaklyWS, laradji2020looc, laradji2019masks}, but their application to remote sensing images is largely underexplored.

Furthermore, existing remote sensing datasets~\cite{sumbul2019bigearthnet, helber2019eurosat, uzkent2019learning} are highly curated to form well-balanced and diversified classes. Simply discarding the labels does not undo this careful selection of examples, which also requires considerable human effort. Our goal is to exploit the massive amount of publicly available remote sensing data for learning good visual representations in a truly unsupervised way. To enable this, we construct a remote sensing dataset from Sentinel-2~\cite{drusch2012sentinel} tiles without any human supervision, neither for curating nor annotating the data.

\begin{figure}[t]
    \begin{center}
    \includegraphics[width=\linewidth, trim=0 100 0 10, clip]{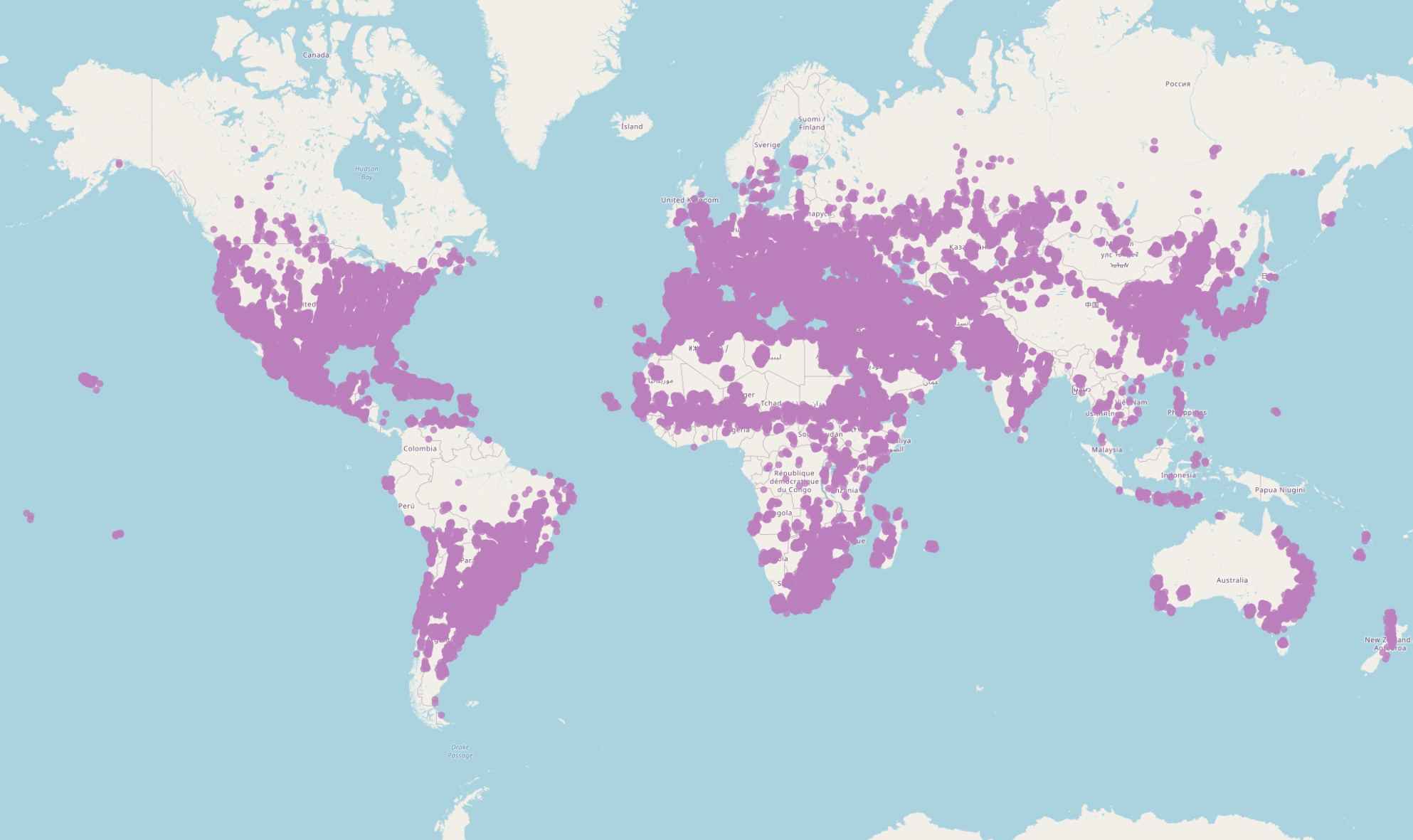}
    \end{center}
    \caption{\textbf{Distribution of the Seasonal Contrast (\methodname{}) dataset.} Each point represents a sampled location. Images are collected around human settlements to avoid monotonous areas such as oceans and deserts.}
    \label{fig:map}
\end{figure}

Another characteristic unique to remote sensing data is satellite revisit, which describes the ability of the system to make repeated image captures of the same point of the Earth's surface over time. For publicly funded satellite constellations such as Sentinel~\cite{drusch2012sentinel} or Landsat~\cite{roy2014landsat}, the revisit time is of the order of days. This temporal dimension provides an additional source of natural variation which complements the artificial augmentation of images. For instance, no amount of artificial augmentation can show how a snowy mountain summit looks like when the snow melts down, or how the different stages of a crop change through the seasons.

Self-supervised learning methods have recently emerged as an effective methodology to learn from vast amounts of unlabeled data. Contrastive methods push together representations of images that are semantically similar (i.e.\ positive pairs). Since no labels are available, traditional contrastive learning methods that work with natural images use different artificial augmentations of the same image (views) as positive pairs. In the case of remote sensing images, we propose to leverage the temporal information to obtain pairs of images from the same location at different points in time, which we call \textit{seasonal positive pairs}. We argue that seasonal changes provide more semantically meaningful content than artificial transformations, and remote sensing images provide this natural augmentation for free. 

We propose \Methodname{} (\methodname{}), a novel methodology for pre-training rich, transferable representations for remote sensing applications. \methodname{} consists of two parts, an unsupervised data acquisition procedure and a self-supervised learning model to learn from the acquired data. The self-supervised learning model is designed based on the observation that encouraging the representation to be invariant to seasonal changes is a strong inductive bias. This property can be beneficial for certain downstream tasks where the prediction will not change with seasonal variations (e.g.\ land-cover classification, agricultural pattern segmentation, building detection), but harmful for downstream tasks where seasonal variations are important (e.g.\ deforestation tracking, change detection). We would like to learn good representations of remote sensing images that are agnostic to the downstream tasks where they could be applied.

To leverage temporal information without limiting the visual representations to be always invariant to time, we use the idea of multiple embedding sub-spaces~\cite{xiao2020should}. Instead of mapping an image to a single embedding space which is invariant to all augmentations, we construct separate embedding sub-spaces and optimize them to be variant or invariant to seasonal changes. We use a multi-head architecture with a shared backbone which produces a common representation that encodes the different variances and invariances. Once the model is trained, this representation can be applied to a wide range of remote sensing downstream tasks, where the model can selectively utilize the different factors of variation captured in the representation.

We evaluate \methodname{} on several remote sensing datasets and tasks. Our experiments on land-cover classification with BigEarthNet~\cite{sumbul2019bigearthnet} and EuroSAT~\citep{helber2019eurosat}, and change detection with OSCD~\citep{daudt2018urban} demonstrate that \methodname{} pre-training is more effective for remote sensing tasks than the common ImageNet~\citep{russakovsky2014imagenet} and MoCo~\citep{he2020momentum} pre-training.

In summary, our contributions are:
\begin{itemize}
    \item We describe a general method for collecting uncurated and unlabeled datasets of remote sensing images. We use this method to construct a remote sensing dataset from Sentinel-2 tiles without any human supervision.
    \item We combine recent contrastive self-supervised learning methods with the temporal information provided by satellites to learn good visual representations which are simultaneously variant and invariant to seasonal changes.
    \item We obtain state-of-the-art results on BigEarthNet and EuroSAT land-cover classification, and on OSCD change detection.
\end{itemize}

\section{Background}
Self-supervised learning is the branch of unsupervised learning where the data itself provides the supervision. The main idea is to occlude or perturb part of the data and task the network with predicting it from the visible data. This defines a pretext task (or proxy loss) and the network is forced to learn what we care about the data (e.g.\ a semantic representation) in order to solve it. A variety of pretext tasks have been proposed for images, such as predicting the relative position of patches~\cite{doersch2015unsupervised}, solving jigsaw puzzles~\cite{noroozi2016unsupervised}, predicting rotations~\cite{gidaris2018unsupervised} or colorization~\cite{zhang2016colorful}.

More recently, contrastive pretext tasks~\cite{wu2018unsupervised, oord2018representation, tian2019contrastive, he2020momentum, misra2020self, chen2020simple, grill2020bootstrap, caron2020unsupervised} have dominated the subfield of self-supervised learning, demonstrating superior performance in various downstream tasks. Intuitively, contrastive learning methods pull together the representations of similar examples while pushing apart the representations of dissimilar examples. Since the examples are not labeled, these methods make the assumption that each example defines and belongs to its own class. Hence, positive pairs are generated by applying random augmentations to the same example, while negative pairs come from other instances in the dataset.

Formally, this task can be formulated as a dictionary look-up problem, where a given example $x$ is augmented into two views, query $x^q$ and key $x^k$, an encoder network $f$ maps the examples into an embedding space, and the representation of the query $q = f(x^q)$ should be closer to the representation of its designated key $k^+ = f(x^k)$ than to the representation of any negative key $k^-$ coming from a set of randomly sampled instances different from $x$. To this end, a contrastive objective is optimized over a batch of positive/negative pairs. A common choice is the InfoNCE loss~\cite{oord2018representation}:

\begin{equation}\label{eq:infonce_loss}
    \mathcal{L} = -\mathrm{log} \frac{\mathrm{exp} (q \cdot k^+ / \tau)}{\mathrm{exp} (q \cdot k^+ / \tau) + \sum_{k^-} \mathrm{exp} (q \cdot k^- / \tau)}
\end{equation}

where $\tau$ is a temperature hyper-parameter scaling the distribution of distances.


\section{Method}
We propose a methodology for pre-training rich, transferable representations for remote sensing imagery, consisting of a general procedure for collecting an unsupervised pre-training dataset (Section~\ref{ssec:dataset_collection}) and a self-supervised learning method (Section~\ref{ssec:seasonal_contrast}) for leveraging this data.

\begin{figure*}[t]
    \begin{center}
    \includegraphics[width=0.9\linewidth]{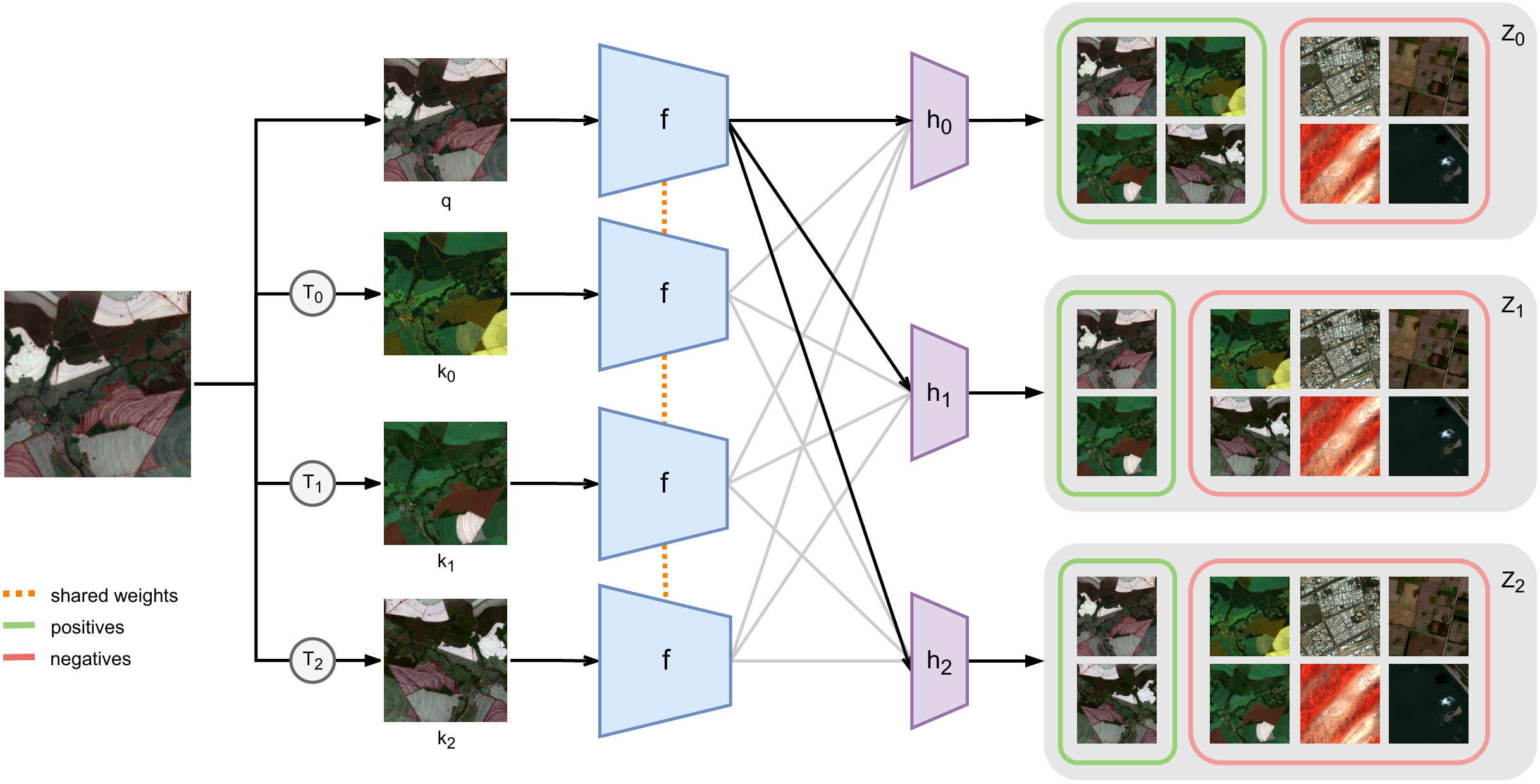}
    \end{center}
    \caption{\textbf{Diagram of the Seasonal Contrast method.} A query image ($q$) is augmented with temporal ($k_{0},k_{1}$) and synthetic ($k_{0},k_2$) transformations $\mathcal{T}$. Image embeddings produced by the encoder $f$ are projected into three different sub-spaces by heads $h_0,h_1,h_2$. Green boxes represent positive pairs while red boxes represent negative pairs (i.e.\ including images from other locations). Sub-space $\mathcal{Z}_0$ is invariant to all transformations, thus all keys belong to the same class as the query. $\mathcal{Z}_1$ is invariant to seasonal augmentations, while $\mathcal{Z}_2$ is invariant to synthetic augmentations.}
    \label{fig:diagram}
\end{figure*}

\subsection{Unsupervised Dataset Collection}
\label{ssec:dataset_collection}
Remote sensing provides a vast amount of imagery data, but annotations are usually scarce, and domain expertise or ground sensors are often required~\cite{jean2016combining}. In order to train on a large amount of satellite images, we collect a new dataset of Sentinel-2~\cite{drusch2012sentinel} patches without any human supervision.

The Sentinel-2 imagery consists of 12 spectral bands (including RGB and NIR) at 10 m, 20 m and 60 m resolution, with a revisit time of around 5 days. We use Google Earth Engine~\cite{gorelick2017google} to process and download image patches from about 200K locations around the world, where each patch covers a region of roughly $2.65 \times 2.65$~km. At each location, we download 5 images from different dates separated by approximately 3 months, which capture the seasonal changes that occurred in the region over a year. To avoid getting images from the same periods of the year, at each location we jitter the dates for up to a year. We also filter out Sentinel-2 tiles with a cloud percentage higher than 10\%. In total, we obtain about 1 million multi-spectral image patches, which amount to a total of over 387 billion pixels.

\vspace{-1em}\paragraph{Sampling Strategy} Our objective is to learn an encoder that can be used on a wide variety of downstream tasks. To this end, we need to sample from a wide variety of regions on the Earth. Uniform sampling would lead to a large amount of redundancy in the types of images. For example, oceans cover 71\% of the planet, forests cover 31\% of land, and deserts cover 33\% of land. To work around this, we make the assumption that most of the variability can be observed in the greater areas around cities. The cities themselves contain a wide range of constructions, a few kilometers away from cities we often observe a variety of crops and industrial facilities. Finally, in the range of 50~km-100~km away from cities, we usually observe natural environments. Hence, we sample around cities following this heuristic (see results in Figure~\ref{fig:map}):

\begin{enumerate}[nolistsep]
    \item Sample uniformly among the 10k most populated cities, and then sample a set of coordinates from the Gaussian distribution spanning a standard deviation of 50~km around the center of the city.
    \item Randomly select a reference date over the past year. Add periodic increments of 3 months to obtain the sampling dates.
    \item For a 15-day range around each date, check if there exists a Sentinel-2 tile with less than 10\% of cloud coverage that intersects with the coordinates.
    \item If there exists a valid Sentinel-2 tile for this location on all dates, process and download all the image patches. Otherwise, go to step 1.
\end{enumerate}

We do not perform any additional data cleaning to ensure that the obtained images are diverse, informative and free of clouds. Because our dataset is constructed automatically, we can easily gather more data (more locations, more dates per location). In this work, however, we limit the scale to a total of 1M images to make it more comparable to ImageNet~\cite{russakovsky2014imagenet}. 

\subsection{Seasonal Contrast}\label{ssec:seasonal_contrast}
Given an unsupervised dataset of remote sensing images with temporal information, we learn a representation that takes advantage of the structure of the data. We get inspiration from~\cite{xiao2020should} to develop a multi-augmentation contrastive learning method. This approach can selectively prevent information loss incurred by artificial augmentations, and extend it with natural augmentations provided by the seasonal changes on remote sensing images. Instead of projecting every view to a common embedding space which is invariant to all augmentations, a common representation is projected into several embedding sub-spaces which are variant or invariant to time (see Figure \ref{fig:diagram}). Hence, the shared representation will contain both time-varying and invariant features, which will transfer efficiently to remote sensing downstream regardless of whether they involve temporal variation.

\subsubsection{Views Generation}
Given a reference image (query), we produce multiple positive pairs (keys) with seasonal and artificial augmentations. Let $\mathcal{T}$ be a set of commonly used artificial augmentations~\citep{he2020momentum}, such as random cropping, color jittering, and random flipping. We first obtain 3 images from the same location at different times, $x^{t_0}$, $x^{t_1}$ and $x^{t_2}$, which are randomly selected among all the available ones for that location. No additional transformations are applied to the query image, i.e.\ $x^q = x^{t_0}$. Hence, $x^{t_1}$ and $x^{t_2}$ can be considered seasonal augmentations (or temporal views) of $x^q$. The first key view contains both seasonal and artificial transformations, $x^{k_0} = \mathcal{T}(x^{t_1})$, the second key view contains only seasonal augmentations, $x^{k_1} = x^{t_2}$, and the third view contains only artificial augmentations, $x^{k_2} = \mathcal{T}(x^{t_0})$.

\subsubsection{Multiple Embedding Sub-spaces}
The query and key views are encoded by a neural network $f$ into representations $v^q$, $v^{k_0}$, $v^{k_1}$, $v^{k_2}$ in a common embedding space $\mathcal{V} \in \mathbb{R}^d$. Next, each intermediate representation is projected into 3 different sub-spaces $\mathcal{Z}_0$, $\mathcal{Z}_1$, $\mathcal{Z}_2 \in \mathbb{R}^{d'}$ by non-linear projection heads $h_0$, $h_1$, $h_2$, where $h_i: \mathcal{V} \mapsto \mathcal{Z}_i$. Following recent literature~\cite{wang2020understanding}, the embedding sub-spaces are $l_2$-normalized, effectively restricting them to the unit hypersphere.

The embedding sub-space $\mathcal{Z}_0$ is invariant to all augmentations, $\mathcal{Z}_1$ is invariant to seasonal augmentations but variant to artificial augmentations, and $\mathcal{Z}_2$ is invariant to artificial augmentations but variant to seasonal augmentations. Namely, in $\mathcal{Z}_0$ all embeddings $z_0^i$ should be pulled together, in $\mathcal{Z}_1$ only $z_1^q$ and $z_1^{k_1}$ should be pulled together and pushed apart from $z_1^{k_0}$ and $z_1^{k_2}$, and in $\mathcal{Z}_2$ only $z_2^q$ and $z_2^{k_2}$ should be pulled together and pushed apart from $z_2^{k_0}$ and $z_2^{k_1}$. This is represented visually in Figure~\ref{fig:diagram}.

A contrastive learning objective is optimized on each embedding sub-space based on Equation \ref{eq:infonce_loss}, where the definition of positive (and negative) pairs depends on the invariances (and variances) that are encoded. In $\mathcal{Z}_0$, the positive pair for the query $z_0^q$ is $z_0^{k_0}$, and the negative pairs are embeddings of other instances in this embedding sub-space. For embedding sub-space $\mathcal{Z}_1$, the positive pair for the query $z_1^q$ is $z_1^{k_1}$, while the negative pairs are embeddings of other instances in this embedding sub-space, plus $z_1^{k_0}$ and $z_1^{k_2}$. Note that $z_1^{k_0}$ and $z_1^{k_2}$ are harder negative pairs for $z_1^q$ as they come from the \textit{same} instance but have a different artificial augmentation. Positive and negative pairs in embedding space $\mathcal{Z}_2$ are analogous to $\mathcal{Z}_1$. 

The final learning objective is the sum of all the embedding sub-space losses. The encoder $f$ must preserve time-varying and invariant information in the general embedding space $\mathcal{V}$ in order to optimize the combined contrastive learning objectives of all normalized embedding sub-spaces $\mathcal{Z}_i$. Note that the original contrastive learning objective~\cite{oord2018representation} is a particular case of multi-augmentation contrastive learning when only the embedding sub-space $\mathcal{Z}_0$ is used.

The representation for transfer learning is taken from the general embedding space $\mathcal{V}$, since we do not assume any \textit{a priori} knowledge about the downstream tasks. In case the right invariances for downstream tasks were known, the representation could be extracted from a particular embedding sub-space $\mathcal{Z}_i$.

\begin{table*}[t]
    \begin{center}
    \resizebox{0.8\linewidth}{!}{
    \begin{tabular}{l|c|cccc|cccc}
        \toprule
        \multirow{3}{*}{Pre-training} & \multirow{3}{*}{Backbone} & \multicolumn{4}{c|}{100k images} & \multicolumn{4}{c}{1M images} \\
        & & \multicolumn{2}{c}{Linear probing} & \multicolumn{2}{c|}{Fine-tuning} & \multicolumn{2}{c}{Linear probing} & \multicolumn{2}{c}{Fine-tuning} \\
        & & 10\% & 100\% & 10\% & 100\% & 10\% & 100\% & 10\% & 100\% \\
        \hline\hline
        Random init. & \multirow{2}{*}{ResNet-18} & 43.05 & 45.95 & 68.11 & 79.80 & 43.05 & 45.95 & 68.11 & 79.80 \\
        ImageNet (sup.) & & 65.69 & 66.40 & 78.76 & 85.90 & 65.69 & 66.40 & 78.76 & 85.90 \\
        \hline
        MoCo-v2 & \multirow{3}{*}{ResNet-18} & 69.70 & 70.90 & 78.76 & 85.17 & 69.28 & 70.79 & 78.33 & 85.23 \\
        MoCo-v2+TP & & 70.20 & 71.08 & 79.80 & 85.71 & 72.58 & 73.60 & 80.68 & 86.59 \\
        SeCo (ours) & & \textbf{74.67} & \textbf{75.52} & \textbf{81.49} & \textbf{87.04} & \textbf{76.05} & \textbf{77.00} & \textbf{81.86} & \textbf{87.27}\\
        \hline\hline
        Random init. & \multirow{2}{*}{ResNet-50} & 43.95 & 46.92 & 69.49 & 78.98 & 43.95 & 46.92 & 69.49 & 78.98 \\
        ImageNet (sup.) & & 70.46 & 71.82 & 80.04 & 86.74 & 70.46 & 71.82 & 80.04 & 86.74 \\
        \hline
        MoCo-v2 & \multirow{3}{*}{ResNet-50} & 71.85 & 73.27 & 79.23 & 85.79 & 73.71 & 75.65 & 80.08 & 86.05 \\
        MoCo-v2+TP & & 72.61 & 73.91 & 79.04 & 85.35 & 74.50 & 76.32 & 80.20 & 86.11 \\
        SeCo (ours) & & \textbf{77.49} & \textbf{79.13} & \textbf{81.72} & \textbf{87.12} & \textbf{78.56} & \textbf{80.35} & \textbf{82.62} & \textbf{87.81} \\
        \bottomrule
    \end{tabular}}
    \end{center}
    \caption{Mean average precision on the BigEarthNet land-cover classification task. Results cover different pre-training approaches and different ResNet backbones. We also explore the effect of the unlabeled pre-training set size between 100k and 1M images, and the size of the BigEarthNet training set between 10\% and 100\%.}
    \label{tab:bigearthnet_results}
    \vspace{-0.5em}
\end{table*}

\section{Experiments}
In this study, we evaluate the learned representations on three downstream tasks: two land-cover classification tasks, where the representation should be invariant to seasonal changes, and a change detection task, where the representation should be variant to seasonal changes.

\vspace{-1em}\paragraph{Pre-training Implementation Details}
We adopt Momentum Contrast (MoCo-v2)~\cite{chen2020improved} as the backbone for our method due to its combination of state-of-the-art performance and memory efficiency. We apply the same artificial augmentations as MoCo-v2, i.e.\ color jittering, random grayscale, Gaussian blur, horizontal flipping, and random-resized cropping. We use a ResNet~\cite{he2016deep} architecture as the feature extractor, and a 2-layer MLP head with a ReLU activation and 128-dimensional output for each embedding sub-space. We also use separate queues~\cite{he2020momentum} for each embedding sub-space, containing 16,384 negative embeddings at a time. We pre-train the network for 200 epochs with a batch size of 256. We use an SGD optimizer with a momentum of 0.9 and a weight decay of 1e-4. We set an initial learning rate of 0.03 and divide it by 10 at 60\% and 80\% of the epochs. A temperature scaling $\tau$ of 0.07 is used in the contrastive loss. Although the collected dataset contains up to 12 spectral bands, in this work we focus on the RGB channels since it is a more general modality.

\vspace{-1em}\paragraph{Methods}
We compare our unsupervised learning approach against several baselines, including random initialization, ImageNet supervised pre-training, and self-supervised pre-training. For the latter, we provide results for MoCo-v2 pre-training on our unsupervised dataset without exploiting the temporal information. In this case, the length of the dataset depends on the total number of images and not the number of geographical locations, so we divide the number of pre-training epochs by the number of images per location. We also provide results for MoCo-v2 pre-training on our dataset leveraging the temporal information for generating positive pairs (MoCo-v2+TP), i.e.\ positive image pairs come from the same location at different times, and MoCo-v2 artificial augmentations are then applied to the spatially aligned image pairs (similar to \citet{ayush2020geography}). We evaluate all methods with linear probing (freezing the encoder and training only the classifier) and fine-tuning (updating the parameters of both the encoder and the classifier).

\subsection{Land-Cover Classification on BigEarthNet}
BigEarthNet~\cite{sumbul2019bigearthnet} is a challenging large-scale multi-spectral dataset of Sentinel-2~\cite{drusch2012sentinel} images, captured with similar sensors to the ones in our unsupervised dataset, i.e.\ 12 frequency channels (including RGB) are provided. It consists of 125 Sentinel-2 tiles acquired between June 2017 and May 2018 over 10 European countries, which are divided into 590,326 non-overlapping image patches, each covering an area of $1.2 \times 1.2$~km with resolutions of 10~m, 20~m, and 60~m per pixel. We discard about 12\% of the patches which are fully covered by seasonal snow, clouds or cloud shadows. This is a multi-label dataset where each image is annotated by multiple land-cover classes, so we measure the downstream performance in terms of mean average precision (mAP). We adopt the new class nomenclature introduced in \cite{sumbul2020bigearthnet}, and we use the same train/val splits proposed in \cite{neumann2019domain}.

\vspace{-1em}\paragraph{Implementation Details}
We evaluate the learned representations by training a linear classification layer with supervised learning. We initialize the ResNet backbone with a pre-trained representation and add a single fully-connected layer which maps from the intermediate representation to class logits. We fine-tune the network for 100 epochs with a batch size of 1024, and report the best validation results for each run. We use an Adam optimizer with default hyper-parameters. For linear probing, we set the initial learning rate to 1e-3; for full fine-tuning, we set the initial learning rate to 1e-5. During training, the learning rate is divided by 10 at 60\% and 80\% of the epochs.

\vspace{-1em}\paragraph{Quantitative Results}
Table~\ref{tab:bigearthnet_results} compares the accuracy of \methodname{} pre-training on BigEarthNet with other pre-training methods. The comparison is done by linear probing or fine-tuning with different backbones, number of pre-training images, and percentage of BigEarthNet labeled data available. For linear probing, we observe that \methodname{} consistently outperforms MoCo-v2+TP. We also observe that temporal positives (TP) improve the performance of MoCo-v2 by a narrow margin. Moreover, we find that \methodname{} features significantly improve over ImageNet pre-trained features, which confirms our hypothesis that there is a gap between remote sensing and natural image domains. We also find that this gap decreases when fine-tuning an ImageNet pre-trained feature extractor on the whole BigEarthNet training set. Nonetheless, with 1M images and a ResNet-50 backbone, \methodname{} features achieve $1.1\%$ higher accuracy than ImageNet features. To the best of our knowledge, this is the first time an unsupervised method obtains higher accuracy than ImageNet pre-training on BigEarthNet with 100\% of the labels. Regarding the backbone size, we observe a wider performance gap between ResNet-18 and ResNet-50 when linear probing than when fine-tuning the whole network. We also find that pre-training with 1M images yields better performance regardless of the backbone used. In all cases, we find that \methodname{} is more efficient than the baselines when only using $10\%$ of BigEarthNet's labeled data; we provide more details in the next section.

\begin{figure}[t]
    \begin{center}
    \includegraphics[width=\linewidth]{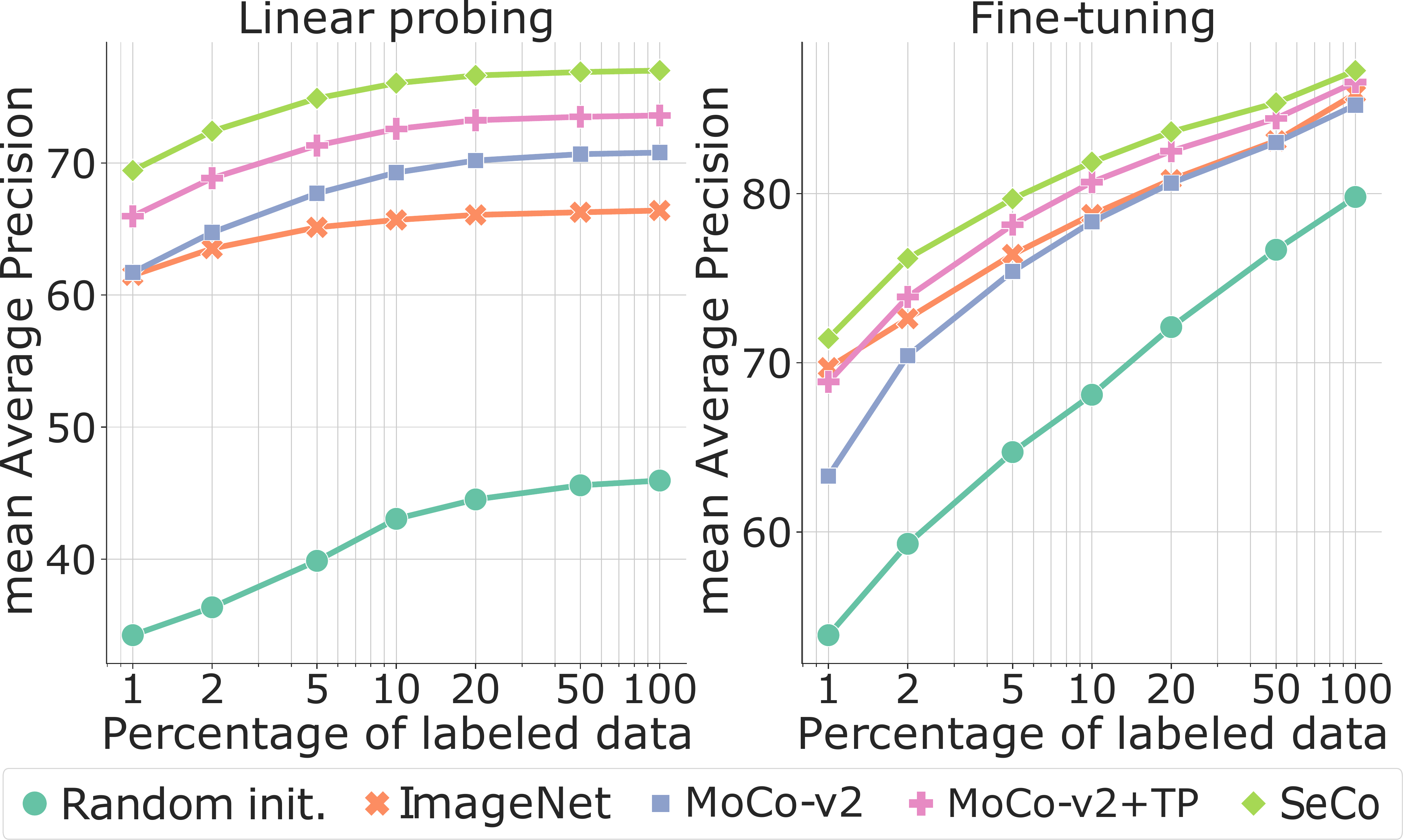}
    \end{center}
    \caption{Label-efficient land-cover classification on BigEarthNet. We use a ResNet-18 backbone pre-trained on 1M images.}
    \label{fig:bigearthnet_results}
\end{figure}

\vspace{-1em}\paragraph{Study on Label-Efficient Transfer Learning}
Figure~\ref{fig:bigearthnet_results} shows the linear probing and fine-tuning performance of \methodname{} and the different baselines for different percentages of labeled data on BigEarthNet. For linear probing, we observe that, with only $1\%$ of the BigEarthNet labels, \methodname{} outperforms ImageNet pre-training with $100\%$ of the labels and matches MoCo-v2 with $20\%$ of the labels. We also observe that the gap between ImageNet pre-training and self-supervised learning increases with the amount of labeled data, while the gap between self-supervised methods does not change significantly. For all percentages of labeled data, \methodname{} achieves a constant $\sim 4\%$ improvement gap with respect to MoCo-v2+TP. When fine-tuning, the performance gap between self-supervised methods and ImageNet narrows down when increasing the percentage of labeled data. Nevertheless, \methodname{} is more label-efficient than all the baselines, matching the performance of ImageNet pre-training using all the available labels with only $50\%$ of the labels.


\vspace{-1em}\paragraph{Ablation on the Locations Sampling Strategy}
In order to evaluate the effectiveness of our sampling strategy to collect uncurated images for pre-training remote sensing representations, we download an alternative version of the \methodname{} dataset where the Earth locations are sampled uniformly from within the continents. We download 100k images following this approach and pre-train a ResNet-18 with the \methodname{} method. Table \ref{tab:ablation_sampling} compares transfer learning performance on the BigEarthNet downstream task when using each sampling scheme. We observe that a \methodname{} representation pre-trained on images sampled from a mixture of Gaussians centered around human settlements (see section \ref{ssec:dataset_collection}) provides better downstream performance than sampling the images uniformly. We argue this is because populated regions tend to be more diverse due to human activity and thus collected images contain more information for learning good representations.

\begin{table}[t]
    \begin{center}
    \begin{tabular}{l|cccc}
        \toprule
        \multirow{2}{*}{Sampling} & \multicolumn{2}{c}{Linear probing} & \multicolumn{2}{c}{Fine-tuning} \\
        & 10\% & 100\% & 10\% & 100\% \\
        \hline\hline
        Gaussian & \textbf{74.67} & \textbf{75.52} & \textbf{81.49} & \textbf{87.04} \\
        Uniform & 71.63 & 72.59 & 79.65 & 85.75 \\
        \bottomrule
    \end{tabular}
    \end{center}
    \caption{Comparison of \methodname{} dataset sampling strategies. We use a ResNet-18 backbone pre-trained on 100k images.}
    \label{tab:ablation_sampling}
\end{table}

\subsection{Land-Cover Classification on EuroSAT}
EuroSAT~\citep{helber2019eurosat} also addresses the challenge of land use and land cover classification using Sentinel-2 satellite images. The images correspond to 34 European countries, and they consist of 10 classes corresponding to different land uses. Each of the classes is composed of 2,000 to 3,000 images, making a total of 27,000 labeled images. The images size is $64 \times 64$~pixels, covering an area of $640 \times 640$~m. All 13 Sentinel-2 spectral bands are included. We adopt the same train/val splits proposed in \cite{neumann2019domain}.

\vspace{-1em}\paragraph{Implementation Details}
On this task, we also evaluate the learned representations by learning a linear classifier with supervised learning. We initialize a ResNet-18 backbone with a pre-trained representation and add a single fully-connected layer on top. In this case, we initialize the backbone with representations pre-trained on 1M satellite images (except when using random weights or loading an ImageNet pre-trained model). We freeze the backbone weights and train the classifier for 100 epochs with a batch size of 32, reporting the best validation accuracy for each run. We use an Adam optimizer with default hyperparameters, setting the initial learning rate to 1e-3 and dividing it by 10 at 60\% and 80\% of the epochs.

\vspace{-1em}\paragraph{Quantitative Results}
Table \ref{tab:eurosat_results} compares the linear probing accuracy of \methodname{} representations against the different baselines. We see that \methodname{} achieves $6.7\%$ higher accuracy than ImageNet pre-training and $3.6\%$ higher accuracy than MoCo-v2+TP. These results confirm that the learned representation is not only effective on BigEarthNet, but also generalizes to other remote sensing datasets such as EuroSAT.


\begin{table}[t]
    \begin{center}
    \begin{tabular}{l|c}
        \toprule
        Pre-training & Accuracy \\
        \hline\hline
        Random init. & 63.21 \\
        Imagenet (sup.) & 86.44 \\
        \hline
        MoCo-v2 & 83.72 \\
        MoCo-v2+TP & 89.51 \\
        SeCo (ours) & \textbf{93.14} \\
        \bottomrule
    \end{tabular}
    \end{center}
    \caption{Fine-tuning accuracy on the EuroSAT land-cover classification task. We use a ResNet-18 backbone pre-trained on 1M images.}
    \label{tab:eurosat_results}
\end{table}

\subsection{Change Detection on Onera Satellite}
The Onera Satellite Change Detection (OSCD) dataset~\citep{daudt2018urban} is composed of 24 pairs of multispectral images from Sentinel-2. The images were recorded between 2015 and 2018 from locations all over the world with various levels of urbanization, where urban changes were visible. Each location contains aligned pairs covering all 13 Sentinel-2 spectral bands. Images vary in spatial resolution between 10~m, 20~m and 60~m, with approximately $600 \times 600$ pixels at 10~m resolution. The goal is to detect changes between satellite images from different dates. Pixel-level change ground truth is provided for all training and validation image pairs. We use the same train/val splits proposed by \citet{daudt2018urban}: 14 images for training and 10 images for validation. We measure the the downstream performance in terms of F1 score, as it is common in the image segmentation literature.

\vspace{-1em}\paragraph{Implementation Details} For every pair of images from a given location at two different timestamps, we produce segmentation masks by following a procedure similar to \citet{daudt2018fully}. First, a ResNet-18 backbone extracts features from each image. We keep the features after each downsampling operation in the backbone network. Then, we compute the absolute value of the difference between the two sets of features in each pair, and use the feature differences as input to a U-Net~\citep{ronneberger2015u} in order to generate binary segmentation masks. The backbone network is initialized with representations pre-trained on 1M satellite images. To avoid overfitting, we freeze the backbone and only train the weights of the U-Net, add a $0.3$ dropout rate after each upsampling layer in the U-Net, and augment the training images with random horizontal flips and $90^{\circ}$ rotations. In addition, since the images in the OSCD dataset have variable size, we split them into non-overlapping patches of $96 \times 96$~pixels. We train the decoder for 100 epochs with a batch size of 32, and report results on the validation set from the point of view of the "change" class. We use an Adam optimizer with a weight decay of 1e-4. We set the initial learning rate to 1e-3 and decrease it exponentially with a multiplicative factor of 0.95 at each epoch.

\vspace{-1em}\paragraph{Quantitative Results}
Table~\ref{tab:oscd_results} compares \methodname{} with random initialization, ImageNet pre-training, MoCO-v2, and MoCo-v2+TP. We observe that \methodname{} initialization achieves higher recall and F1 score than all the baselines. In particular, \methodname{} outperforms MoCo-v2+TP by 6.8\% F1 score. This might be due to MoCo-v2+TP representations being invariant to temporal variations, which is not a desirable property in a change detection task. Interestingly, although both \methodname{} and MoCo-v2 consider image patches from the same location at different timestamps as negative pairs (i.e.\ their learned representations are variant to time), \methodname{} attains a 6.2\% higher F1 score. This indicates that the multiple embedding sub-spaces make \methodname{} more effective at detecting temporal changes by disentangling image augmentations from temporal variations.


\begin{table}[t]
    \begin{center}
    \begin{tabular}{l|ccc}
        \toprule
        Pre-training & Precision & Recall & F1 \\
        \hline\hline
        Random init. & \textbf{70.53} & 19.17 & 29.44 \\
        Imagenet (sup.) & \textbf{70.42} & 25.12 & 36.20 \\
        \hline
        MoCo-v2 & 64.49 & 30.94 & 40.71 \\
        MoCo-v2+TP & 69.14 & 29.66 & 40.12 \\
        SeCo (ours) & 65.47 & \textbf{38.06} & \textbf{46.94} \\
        \bottomrule
    \end{tabular}
    \end{center}
    \caption{Fine-tuning results on the Onera Satellite change detection task. We use a ResNet-18 pre-trained on 1M images.}
    \label{tab:oscd_results}
\end{table}

\begin{figure*}[t]
    \begin{center}
    \includegraphics[width=\linewidth]{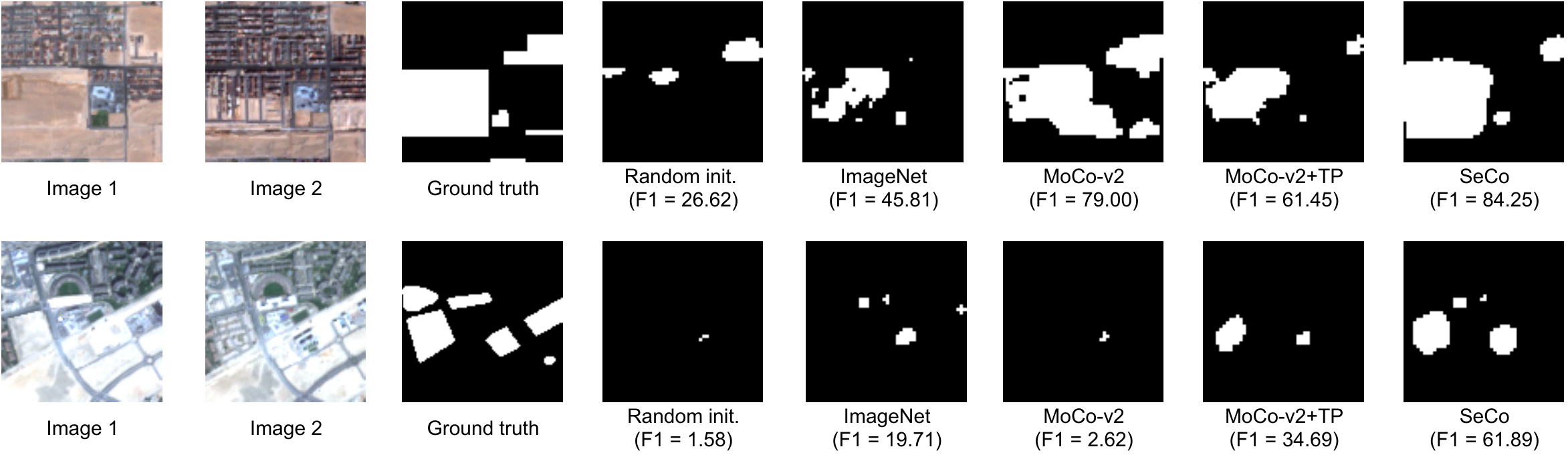}
    \end{center}
    \caption{Comparison of qualitative results on the Onera Satellite change detection task. Each row contains the input images, the ground truth mask, and the generated change detection masks for a validation sample.}
    \label{fig:oscd_results}
    \vspace{-0.5em}
\end{figure*}

\vspace{-1em}\paragraph{Qualitative Results}
Figure \ref{fig:oscd_results} compares the change detection masks produced by our method and all the baselines on two samples from the OSCD validation set. We observe that \methodname{} pre-training produces higher quality masks which cover more of the changed pixels without excessive false negatives. We also notice some discrepancies in the performance of MoCo-v2 with and without leveraging temporal information (TP). We hypothesize these might be due to the different treatment of temporal invariance by each approach, and the image differences resembling more artificial augmentations or temporal changes. \methodname{} overcomes this problem by learning a representation that preserves time-varying and invariant factors.

\section{Related Work}

\paragraph{Learning from Uncurated Data}
Recent efforts in unsupervised feature learning have focused on either small or highly curated datasets like ImageNet, whereas using uncurated raw datasets was found to decrease the feature quality when evaluated on a transfer task~\cite{doersch2015unsupervised, caron2018deep}. \citet{caron2019unsupervised} propose a self-supervised approach which leverages clustering to improve the performance of unsupervised methods trained on uncurated data. Other methods use metadata such as hashtags~\cite{joulin2016learning, sun2017revisiting, mahajan2018exploring}, geolocation~\cite{weyand2016planet} or the video structure~\cite{gordon2020watching} as a source of noisy supervision. In our work, we leverage the geographical and temporal information of remote sensing data to learn unsupervised representations from uncurated datasets.

\vspace{-1em}\paragraph{Multi-augmentation Contrastive Learning}
Recent self-supervised contrastive learning methods have been able to produce impressive transferable visual representations by learning to be invariant to different image augmentations. However, these methods implicitly assume a particular set of representational invariances, and can perform poorly when a downstream task violates this assumption. \citet{xiao2020should} propose Leave-one-out Contrastive Learning (LooC), a multi-augmentation contrastive learning framework that produces visual representations able to capture varying and invariant factors by constructing separate embedding spaces, each of which is invariant to all but one augmentation. In our work, we use a similar approach to learn representations that are variant and invariant to the seasonal changes present in remote sensing images.

\vspace{-1em}\paragraph{Unsupervised Learning in Remote Sensing}
While unsupervised learning has been extensively studied on natural image datasets (e.g. ImageNet), this subfield remains underexplored on the remote sensing domain. This is quite surprising given the importance of remote sensing for Earth observation, the vast amount of readily available data, and the many opportunities for self-supervision from the unique characteristics of satellite images. For instance, \citet{jean2019tile2vec} use the geographical information of images to sample positive and negative pairs and build a pretext task based on the triplet loss. \citet{uzkent2019learning} pair georeferenced Wikipedia articles with satellite images of the corresponding locations, and learn representations by predicting properties of the articles from the images. \citet{vincenzi2020color} leverage the multi-spectrality of remote sensing images to build a colorization task, where they reconstruct the visible colors from the other spectral bands. More similar to our work, \citet{ayush2020geography} also propose to exploit the temporal information in satellite imagery to generate positive pairs and train a contrastive objective. However, their representations are always invariant to temporal changes, which might be detrimental for downstream tasks involving temporal variation. We overcome this problem by using multi-augmentation contrastive learning, where the representations preserve time-varying and invariant information.

\section{Conclusions}
We presented \Methodname{} (\methodname{}), a new transfer learning pipeline for remote sensing imagery. \methodname{} consists of a data collection strategy and a self-supervised learning algorithm that leverages this data. First, we sample locations around populated regions over multiple timestamps, which provides a diverse set of satellite images. Then, we extend multi-augmentation contrastive learning methods to take into account the seasonal changes and learn rich and transferable remote sensing representations.

We compared \methodname{} with the common ImageNet pre-training and MoCo pre-training on the collected data using different backbones and dataset sizes. We found that \methodname{} outperforms the considered baselines on BigEarthNet, EuroSAT and OSCD tasks. Thus, we conclude that domain-specific unsupervised pre-training is more effective for remote sensing applications than pre-training with standard datasets such as ImageNet or algorithms such as MoCo.


{\small
\bibliographystyle{abbrvnat}
\bibliography{references}
}

\end{document}